\documentstyle{article}
\catcode`@=11
\def\@bindentby#1{%
\begingroup\setbox3=\hbox{#1}\ifhmode\par\fi\noindent%
\dimen2=\linewidth\dimen4=\@totalleftmargin%
\copy3\unskip\advance\linewidth-\wd3\advance\@totalleftmargin\wd3%
\parshape=2\dimen4\dimen2\@totalleftmargin\linewidth%
\everypar{\parshape=1\@totalleftmargin\linewidth}%
\bgroup\ignorespaces}

\def\@eindentby{%
\egroup\advance\linewidth\wd3%
\advance\@totalleftmargin-\wd3%
\everypar{\parshape=1\@totalleftmargin\linewidth}%
\par\endgroup}

\def\fixindent{%
\egroup\everypar{\parshape=1\@totalleftmargin\linewidth}\par\bgroup%
}

\catcode`@=12

\catcode`@=11
\def\@tabsize{20pt}

\newcount\@lkind
\newcount\@icount
\newcount\@lcount
\newcount\@inmerge
\newcount\@algcount

\newdimen\@tabmargin
\newcounter{@glcount}

\newif\if@algmerged

\global\@lkind	   = 0
\global\@algcount  = 0
\global\@tabmargin = 0pt

\def\@noalgorithm{
	\bgroup\def\@alglabel{}\def\@algcurlabel{}\def\@algsep{}\def\@algprefix{}%
	\@lkind=0\@algmergedfalse\medskip\par
}

\def\@enoalgo{
	\egroup\everypar{\parshape=1\@totalleftmargin\linewidth\noindent}
}

\def\@enoalgorithm{
	\@enoalgo\medskip
}

\def\@balgorithm#1#2{
	\par\global\advance\@algcount1\edef\@currentlabel{\number\@algcount}\setbox5=\hbox{\rm #1}
	\medskip\noindent{{\bf Algorithm} \number\@algcount : \copy5(\hbox{#2})}\@noalgorithm
}

\def\@ealgorithm{\@enoalgo\vskip1.5pt\par\noindent{\bf End \box5}\medskip}

\def\@bsubroutine#1#2{
	\par\setbox5=\hbox{\rm #1}\medskip\noindent{{\bf Subroutine} : \copy5(\hbox{#2})}
	\bgroup\def\@alglabel{}\def\@algcurlabel{}\def\@algsep{}\def\@algprefix{}
	\@lkind=0\@algmergedfalse\vskip1.5pt\par
}

\def\@bindentmerged{
	\begingroup\@icount=0\let\item=\@iitem\let\noitem=\@nitem\let\mitem=\@mitem\@inmerge=1
	\edef\@algprefix{\@algcurlabel}\@lcount=\value{@glcount}\setcounter{@glcount}{0}
}

\def\@bindent{
	\begingroup\@icount=0\let\item=\@iitem\let\noitem=\@nitem\let\mitem=\@mitem\@inmerge=0
	\advance\@tabmargin\@tabsize\advance\linewidth-\@tabsize\advance\@totalleftmargin\@tabsize
	\everypar{\parshape=1\@totalleftmargin\linewidth\noindent}
}

\def\@eindent{
	\ifnum0<\@icount\@eindentby\fi\ifnum\@inmerge=0
	\ifdim\@tabmargin>0pt\advance\@tabmargin-\@tabsize\fi
	\advance\linewidth\@tabsize\advance\@totalleftmargin-\@tabsize
	\everypar{\parshape=1\@totalleftmargin\linewidth\noindent}
	\fi\ifnum\@lcount>-1\setcounter{@glcount}{\@lcount}\fi\endgroup
}

\def\@bnonblank{
	\@lcount=\value{@glcount}\setcounter{@glcount}{0}\@tabmargin=0pt\edef\@algprefix{}

}

\def\@listfill#1{\ifdim\@tabmargin>0pt\hbox{\kern-1.5em\kern-\@tabmargin\hbox to1.5em{#1\hfil}%
	\unskip\kern\@tabmargin}\else\hbox to1.5em{#1\hfil}\fi
}

\def\@listbl{
	\@bindent\@lcount=-1\@algmergedfalse
}

\def\@listar{
	\ifnum\@lkind=1\def\@algsep{.}\fi\edef\@alglabel{\@alglabel\@algcurlabel\@algsep}
	\@lkind=1\def\@algsep{}\if@algmerged\@bindentmerged%\@algmergedfalse
	\else
	\@bindent\@bnonblank\fi
}

\def\@listlc{
	\edef\@alglabel{\@alglabel\@algcurlabel\@algsep}\@lkind=2\def\@algsep{}
	\if@algmerged\@bindentmerged%\@algmergedfalse
	\else\@bindent\@bnonblank\fi
}

\def\@listuc{
	\edef\@alglabel{\@alglabel\@algcurlabel\@algsep}\@lkind=3\def\@algsep{}
	\if@algmerged\@bindentmerged%\@algmergedfalse
	\else\@bindent\@bnonblank\fi
}

\def\@arlabel{
	\def\@algcurlabel{\arabic{@glcount}}\def\@algusedlabel{\@algprefix\@algcurlabel}
	\xdef\@currentlabel{\@alglabel\@algsep\@algcurlabel}
}

\def\@lclabel{
	\def\@algcurlabel{\alph{@glcount}}\def\@algusedlabel{\@algprefix\@algcurlabel}
	\xdef\@currentlabel{\@alglabel\@algsep\@algcurlabel}
}

\def\@uclabel{
	\def\@algcurlabel{\Alph{@glcount}}\def\@algusedlabel{\@algprefix\@algcurlabel}
	\xdef\@currentlabel{\@alglabel\@algsep\@algcurlabel}
}

\def\@aritem{
	\@arlabel\@bindentby{\@listfill{\@algusedlabel.}}
}

\def\@lcitem{
	\@lclabel\@bindentby{\@listfill{\@algusedlabel.}}
}

\def\@ucitem{
	\@uclabel\@bindentby{\@listfill{\@algusedlabel.}}
}

\def\@blitem{
	\@bindentby{\hbox to 0em{\hfil}}
}

\def\@iitem{
	\if@algmerged\@inmerge=0\@algmergedfalse\unskip\else
	\ifnum0<\@icount\unskip\@eindentby\fi\fi
	\advance\@icount1\global\stepcounter{@glcount}
	\ifcase\@lkind\@blitem\or\@aritem\or\@lcitem\or\@ucitem\fi\ignorespaces\unskip
}

\def\@nitem{
	\ifnum0<\@icount\unskip\@eindentby\fi\advance\@icount1\@bindentby{\@listfill{\hbox{}}}%
	\ignorespaces\unskip
}

\def\@mitem{
	\ifnum0<\@icount\unskip\@eindentby\fi\advance\@icount1\global\stepcounter{@glcount}
	\ifcase\@lkind\@bllabel\or\@arlabel\or\@lclabel\or\@uclabel\fi
	\@algmergedtrue\ignorespaces\unskip
}

\newenvironment{algorithm}[1]{\@balgorithm{#1}}{\@ealgorithm}

\newenvironment{listbl}{\@listbl}{\@eindent}

\catcode`@=12

\begin{document} 
\title{Domain Adaptation with Clustered Language Models
\footnote{Preprint - To Appear in ICASSP97}
}
\author{J.P. Ueberla\\
Forum Technology - DRA Malvern, St.Andrews Road\\
Malvern, Worcestershire, WR14 3PS, UK\\
email:ueberla@signal.dra.hmg.gb}

\maketitle

\begin{abstract}
In this paper, a method of domain adaptation for clustered language models is developed. It is based on a previously developed clustering algorithm, but with a modified optimisation criterion. The results are shown to be slightly superior to the previously published 'Fillup' method, which can be used to adapt standard n-gram models. However, the improvement both methods give compared to models built from scratch on the adaptation data is quite small (less than 11\% relative improvement in word error rate). This suggests that both methods are  still unsatisfactory from a practical point of view.
\end{abstract}

\section{Introduction}

Current large vocabulary speech recognition systems can  achieve good performance on domains for which large quantities (e.g. millions of words) of textual data are available to train a language model. In real world applications, however, this  is quite often not the case. The issue of language model domain adaptation is therefore of great practical importance. 

One approach to tackle this problem is to try to learn from an analogy to the speaker dependence issue: current systems perform well by training speaker independent models, which can then be adapted with relatively little data from a given speaker (see \cite{Leg95}). Can the same approach be applied to language model adaptation? 

In section \ref{background}, previous work in this area is reviewed and 
a rough working definition of domain is given. A  method to perform domain adaptation with clustered language models is then developed (Section \ref{adaptive}). Experimental results to evaluate the method are given in Section \ref{results}, followed by conclusions in Section \ref{conclusion}.

\section{Background}
\label{background}

In order to make the description of domain adaptation more precise, a definition of domain is needed. One might be tempted to define domain in the sense of semantic topic.
However, texts might differ  in other aspects (e.g. style), which
could still require language model adaptation. A more general definition of domain, more in line with the term sublanguage, is therefore required. According to \cite{McN92}, there are many
different definitions of the term, but most of them seem to agree on the
following characteristics of a sublanguage:
\begin{enumerate}
\item it is part of a natural language
\item it is of a specialised form
\item it behaves like a complete language
\item it is used in special circumstances (e.g. expert communication)
\item it is limited to a particular subject domain
\end{enumerate}
Some of these points seem very useful for the concept of domain (2,4),
others less so (1). What properties should an acceptable definition of 
domain have? The following spring to mind:
\begin{itemize}
\item there should be a continuum  (e.g. an infinite number) of domains
\item each domain may contain an infinite number of elements (e.g. 
documents/sentences/words)
\item for a given element, one should be able to decide whether or not it 
belongs to a given domain
\item all elements of a domain should have a common feature (which defines 
the domain)
\end{itemize}
This leads to the following rather wide working definition of domain and 
hence domain adaptation: A {\bf domain} D is a (often infinite) set of documents such that each document satisfies a property $P_{D}$ (e.g. 'the document deals with some aspect of law'). Given a sample $S_{Back}$ of domain $D_{Back}$ (background domain) and a sample $S_{Adapt}$ of domain $D_{Adapt}$ (target domain), the problem of language model {\bf domain adaptation} is to produce a language model for $D_{Adapt}$ by using $S_{Adapt}$ and by carrying over some of the information contained in $S_{Back}$.

Domain adaptation can be  divided into static and dynamic domain 
adaptation,
depending on the time scale used to perform adaptation. Dynamic adaptation 
tries to capture phenomena with a shorter time scale (e.g. topic shifts) and
is performed on line,
whereas static adaptation can be used to perform a one-time shift from one
domain to another and is performed off line. Previous work has shown
improvements by using both dynamic adaptation of n-gram models (\cite{Kuh90}, 
\cite{Jel91}, \cite{Pie92}, \cite{Mat92}, \cite{Kne93b}, \cite{Ros96}, \cite{Fed96}, \cite{Iye96}) and by using static adaptation of n-gram models (\cite{Mat92}, \cite{Kne93b}, \cite{Bes95}, \cite{Ros96}). Since the 'Fillup' method
presented in \cite{Bes95} gives better performance than linear interpolation,
the 'Fillup' method is used as method of comparison for the adaptive clustering,
which will be developed in the next section.

\section{Adaptive Clustering}
\label{adaptive}

The task of a language model is to calculate $p(w_{i}|c_{i})$, the probability of the next word being $w_{i}$ given the current context $c_{i}$. Language models differ in the way this probability is modelled and how the context $c_{i}$ is defined. A quite general model proposed in \cite{Ney93b} makes use of a state mapping function $S$ and a category mapping function $G$. The idea behind the state mapping $S:c->s_{c}=S(c)$ is to assign each of the large number of possible contexts $c \in C$ to one of a smaller number of context-equivalent states. Similarly, the category mapping $G:w->g_{w}=G(w)$ assigns each of the large number of possible words $w \in V$ to one of a smaller number of categories (similar to parts of speech). The probability of the next word is then calculated as
\begin{equation} 
\label{eq:model}
p(w_{i}|c_{i})=p(G(w_{i})|S(c_{i}))*p(w_{i}|G(w_{i})) . 
\end{equation}
In order to determine $S$ and $G$ automatically, a clustering algorithm as shown in Figure \ref{fig:algo_old} can be used.
\begin{figure}[h]
\begin{algorithm}{Clustering}{}
\begin{listbl}
	\item  start with initial clustering functions $S$, $G$
	\item iterate until some convergence criterion is met
	\begin{listbl}
		\item for all $w \in V$ and $c \in C$
		\begin{listbl}
			\item for all $g'_{w} \in G$ and $s'_{c} \in S$
			\begin{listbl}
				\item calculate the difference in the optimisation criterion when $w$/$c$ is moved from $g_{w}$/$s_{c}$ to $g'_{w}$/$s'_{c}$
			\end{listbl}
			\item move the $w$/$c$ to the $g'_{w}$/$s'_{c}$ that results in the biggest improvement in optimisation criterion
		\end{listbl}
	\end{listbl}
\end{listbl}
\end{algorithm}
\caption{The clustering algorithm}
\label{fig:algo_old}
\end{figure}
It is a greedy, hill-climbing algorithm that moves elements to the best available choice at any given time. 
Based on equation \ref{eq:model} and on the leaving-one-out likelihood of the model generating the training data, an optimisation criterion can be derived (see \cite{Ueb94e} for a detailed description). Let $N(e)$ denote the number of times event $e$ appeared in the training data, let $B$ denote the smoothing parameter used for absolute discounting (\cite{Ney93}), and let $n_{0}, n_{1}, n_{+}$ denote the number of pairs $(s,g)$ that have appeared zero, one and one or more times in the training data. The resulting optimisation criterion $F$ (as derived in \cite{Ueb94e}) is
\begin{eqnarray}
\label{eq:normal}
F & = & \sum_{s,g:N(s,g)>1} N(s,g) * log ( N(s,g) - 1 - B )\\
 & + & n_{1} * log ( \frac{B*(n_{+}-1)}{(n_{0}+1)} ) \nonumber \\
 & - & \sum_{s} N(s) * log (  N(s) - 1 ) - \sum_{g} N(g) * log (  N(g) - 1 )
\nonumber .
\end{eqnarray}

The basic building block in the derivation of equation \ref{eq:normal} is the likelihood of one event in the training corpus, as estimated from the training corpus in which this one event has been removed (leaving-one-out likelihood). The main idea behind the adaptive clustering is to use as basic building block the likelihood of one event in $S_{Adapt}$, as estimated from a linear interpolation of counts from $S_{Back}$ and $S_{Adapt}$ from which this one event has been removed. The motivation for this is that the clustering can thus optimise the perplexity on $S_{Adapt}$, while having access to a linear combination of counts from $S_{Back}$ and $S_{Adapt}$.

Let $N_{A}(e)$ ($N_{B}(e)$) denote the number of times event $e$ appeared in $S_{Adapt}$ ($S_{Back}$). Define $N_{C}(e)$ to be the linear interpolation of the two counts
\begin{equation}
N_{C}(e)=Round(\lambda*N_{A}(e)+(1-\lambda)*N_{B}(e))
\end{equation}
where $Round(x)$ returns the integer nearest to $x$. 
The only events that can contribute to the optimisation function are events that occur at least once in $S_{Adapt}$ (because, as explained above, the likelihood of $S_{Adapt}$ is taken as optimisation function). However, their probability is calculated based on the combined counts. Therefore, the smoothing has to apply to the combined counts. Define
$n_{bi,0},n_{bi,1},n_{bi,+}$ as the number of pairs $(s,g)$ that have a combined count $N_{C}(s,g)$ of $0$, of $1$, and larger than $0$.
In order to introduce absolute discounting for the unigram estimates as well, also define $n_{s,0},n_{s,1},n_{s,+}$ as the number of states $s$ that have a combined count $N_{C}(s)$ of $0$, of $1$, and larger than $0$ (similarly, define  $n_{g,0}$ etc.  for the unigram estimates involving $g$). 
Changing equation \ref{eq:normal} according to the basic idea outlined above, this leads to
\begin{eqnarray}
\label{eq:adaptive}
\lefteqn{F_{adapt}=} \\
&  \sum_{s,g:N_{A}(s,g)>=1,N_{C}(s,g)>1} N_{A}(s,g) * log ( N_{C}(s,g) - 1 - B ) \nonumber & \\
&  +n_{bi,1} * log ( \frac{B*(n_{bi,+}-1)}{(n_{bi,0}+1)} )  \nonumber \\
&  -\sum_{s:N_{A}(s)>=1,N_{C}(s)>1} N_{A}(s) * log (  N_{C}(s) - 1 -B )  \nonumber \\
&  -\sum_{g:N_{A}(g)>=1,N_{C}(g)>1} N_{A}(g) * log (  N_{C}(g) - 1 -B )  \nonumber \\
&  -n_{s,1}*log ( \frac{B*(n_{s,+}-1)}{(n_{s,0}+1)} ) - n_{g,1}*log ( \frac{B*(n_{g,+}-1)}{(n_{g,0}+1)} )
\nonumber .
\end{eqnarray}

By using the same clustering algorithm as before, but with $F_{Adapt}$ instead of $F$ as optimisation criterion, language model domain adaptation can be performed.

\section{Results}
\label{results}

In order to test different adaptation methods, two textual samples $S_{Back}$ and $S_{Adapt}$ and acoustic testing data from $D_{Adapt}$ are required. Since the WSJ domain has the associated acoustic data, it is used as $D_{Adapt}$. As $D_{Back}$, the patent domain (PAT) was chosen, for which a large sample $S_{Adapt}$ (about 35 million words are used) is also available from the LDC as part of the TIPSTER database.

The recognition system is a state-of-the-art HMM based system (continuous densities, mixtures, triphones). All experiments are based on bigram language models, either clustered (500 clusters) or backoff (singleton bigrams were ignored). The different methods evaluated were
\begin{itemize}
\item $Back_{Bo}$: a backoff model built on the background corpus
\item $Back_{Cl}$: a clustered model built on the background corpus
\item $Adapt_{Cl}$: a clustered model built on the adaptation data
\item $Adapt_{Bo}$: a backoff model built on the adaptation data
\item $Fillup$: a model built according to the 'Fillup' method presented in \cite{Bes95}
\item $ClustAdapt$: a model built with the adaptive clustering presented in the previous section; the initial starting point for the clustering is taken to be the clustering produced by $Back_{Cl}$; one global $\lambda$ parameter was used and optimised iteratively at the end of each iteration;
\end{itemize}

For all methods except $Back_{Bo}$ and $Back_{Cl}$, the vocabulary was defined to be all the words that appeared in $S_{Adapt}$, plus additional words from $S_{Back}$ until 20K words were reached. For $Back_{Bo}$ and $Back_{Cl}$, the vocabulary consisted of the 20K most frequent words in $S_{Back}$. Because of this difference, the perplexities of $Back_{Bo}$ and $Back_{Cl}$ are not directly comparable to those of the other models. For each method and a given amount of adaptation material, the perplexity of the resulting model was calculated on a held-out section of $S_{Adapt}$ and a recognition run was performed on the acoustic data.

Table \ref{tab:baseline} gives the results of the two baseline methods, which do not use any of the adaptation material.
\begin{table}
\centering
\begin{tabular}{|c|c|c|} \hline
Model & PP  & WER (\%) \\  \hline
$Back_{Cl}$	&955 	& 49.3 \\
$Back_{Bo}$	&954 	& 48.4 \\ \hline
\end{tabular}
\caption{Baseline results}
\label{tab:baseline}
\end{table} 
The high perplexities show that the PAT and WSJ domains are considerably different. The rate of out-of-vocabulary words is about 15\%, which is one reason
for the very high error rate.

Tables \ref{tab:adaptbo}, \ref{tab:adaptclust},  \ref{tab:fillup} and \ref{tab:adaptiveclust} give the results for the different methods and different amounts of adaptation material.

\begin{table}
\centering
\begin{tabular}{|c|c|c|} \hline
Adapt. words & PP  & WER (\%)\\  \hline
200	& 6130	& 57.0\\
1000	& 2740	& 54.0\\
5000	& 1740	& 47.6\\
25000	& 966	& 39.2\\
125000	& 593	& 33.0\\ \hline
\end{tabular}
\caption{Results for $Adapt_{Bo}$}
\label{tab:adaptbo}
\end{table} 

Comparing Table \ref{tab:adaptclust} to Table \ref{tab:adaptbo}, one can seen that $Adapt_{Cl}$ is more robust than $Adapt_{Bo}$ and it leads to better recognition results for almost the entire range of adaptation material. This is consistent with
previous results (see \cite{Ueb95b}), which showed that clustered models are more robust in terms of perplexity. 

\begin{table}
\centering
\begin{tabular}{|c|c|c|} \hline
Adapt. words & PP  & WER (\%)\\  \hline
200	& 4170	& 57.0\\
1000	& 2150	& 51.1\\
5000	& 1210	& 46.4\\
25000	& 765	& 37.0\\
125000	& 498	& 33.4\\ \hline
\end{tabular}
\caption{Results for $Adapt_{Cl}$}
\label{tab:adaptclust}
\end{table} 

Comparing Table \ref{tab:fillup} to Table \ref{tab:adaptclust}, one can see that $Fillup$ outperforms $Adapt_{Cl}$ in almost all cases.

\begin{table}
\centering
\begin{tabular}{|c|c|c|} \hline
Adapt. words & PP  & WER (\%)\\  \hline
200	& 848	& 49.9\\
1000	& 772	& 49.6\\
5000	& 628	& 44.9\\
25000	& 543	& 38.2\\
125000	& 420	& 33.0\\ \hline
\end{tabular}
\caption{Results for $Fillup$}
\label{tab:fillup}
\end{table} 

By looking at Table \ref{tab:adaptiveclust}, one can see that $ClustAdapt$ outperforms $Fillup$ in almost all cases.

\begin{table}
\centering
\begin{tabular}{|c|c|c|} \hline
Adapt. words & PP  & WER (\%)\\  \hline
200	& 941	& 50.9\\
1000	& 1040	& 48.1\\
5000	& 821	& 44.8\\
25000	& 801	& 38.0\\
125000	& 623	& 32.8\\ \hline
\end{tabular}
\caption{Results for $ClustAdapt$}
\label{tab:adaptiveclust}
\end{table} 

Finally, When comparing table \ref{tab:adaptiveclust} to table \ref{tab:adaptclust}, one can see that the relative
improvements in word error rate by using $ClustAdapt$ instead of $Adapt_{Cl}$ are 10.7\%, 5.87\%, 3.45\%, -2.70\% and 1.80\%.

\section{Conclusion}
\label{conclusion}

Compared to the success of some methods for acoustic adaptation, the results
obtained here are somewhat disappointing. In particular, they seem
to suggest that the improvements from the adaptation techniques compared
to starting from scratch on the adaptation data become quite
small when several tens of thousands of words are available 
\footnote{However, it is important to note that this threshold will 
depend on how dissimilar the two domains are. Moreover, a more fine grained
analysis for different amounts of adaptation data would be beneficial, especially since the results for 25,000 words seem to be falling somewhat outside
the trend.}.
One reason for this could be the
fact that the acoustic space has an underlying distance metric and thus
allows the comparison of two elements. Moreover, one can specify the kind of transformations
one would want the adaptation to be able to perform. Both of these points seem
more difficult in the case of language model adaptation.

Even though the adaptation method for clustered language models developed in this paper gives slightly better results than the 'Fillup' method, the accuracies
obtained with the adaptive clustering and the 'Fillup'  method are still very low compared to the about 80\% or more the system can achieve with a backoff bigram trained on about 40 million words of the WSJ corpus. Both adaptation methods are therefore still unsatisfactory from a practical point of view.

%\nocite{*}
%\bibliography{/home/ueberla/latex/inputs/all,/home/ueberla/latex/inputs/alphabetical}
%\bibliography{auth_kit}
\bibliographystyle{plain}

\end{document}